\documentclass[electronic]{vgtc}             % electronic version

\graphicspath{{figures/}{pictures/}{images/}{./}} % where to search for the images

\usepackage{times}                     % we use Times as the main font
\usepackage{tabu}                      % only used for the table example
\usepackage{booktabs}                  % only used for the table example
\usepackage{lipsum}                    % used to generate placeholder text
\usepackage{mwe}                       % used to generate placeholder figures
\usepackage{tabularx}
\usepackage{siunitx}
\usepackage{mathptmx}                  % use matching math font
\usepackage{subcaption}
\usepackage{xcolor}
\usepackage{colortbl}
\usepackage{multirow}
\usepackage{amsmath}
\usepackage{amssymb}
\usepackage{svg}
\usepackage{threeparttable}
\usepackage{tikz}
\usetikzlibrary{fit,arrows.meta}

\newcommand{\ie}{\textit{i}.\textit{e}., }
\newcommand{\eg}{\textit{e}. \textit{g}., }

\onlineid{2060}

\vgtccategory{Research}

\vgtcinsertpkg

\title{MuCHMDPoser: Multiview RGB-Camera Full-Body Pose Tracking for Virtual Reality}

\author{Antoine Maoirca\thanks{e-mail: antoine.maiorca@umons.ac.be}\\ %
        \parbox{1.4in}{\scriptsize \centering ISIA Lab \\ UMONS \\ Mons, Belgium}
\and George Fletcher \thanks{e-mail:gf321@bath.ac.uk}\\ %
     \parbox{1.4in}{\scriptsize \centering Department of Computer Science \\ University of Bath \\Bath, UK}\\%
\and Thierry Ravet\thanks{e-mail: thierry.ravet@umons.ac.be}\\ %
        \parbox{1.4in}{\scriptsize \centering ISIA Lab \\ UMONS \\ Mons, Belgium}
\and Thierry Dutoit\thanks{e-mail: thierry.dutoit@umons.ac.be}\\ %
        \parbox{1.4in}{\scriptsize \centering ISIA Lab \\ UMONS \\ Mons, Belgium}
}

\teaser{
  \centering
    \hspace*{-10pt} % adjust value as needed
  \includegraphics[width=\linewidth]{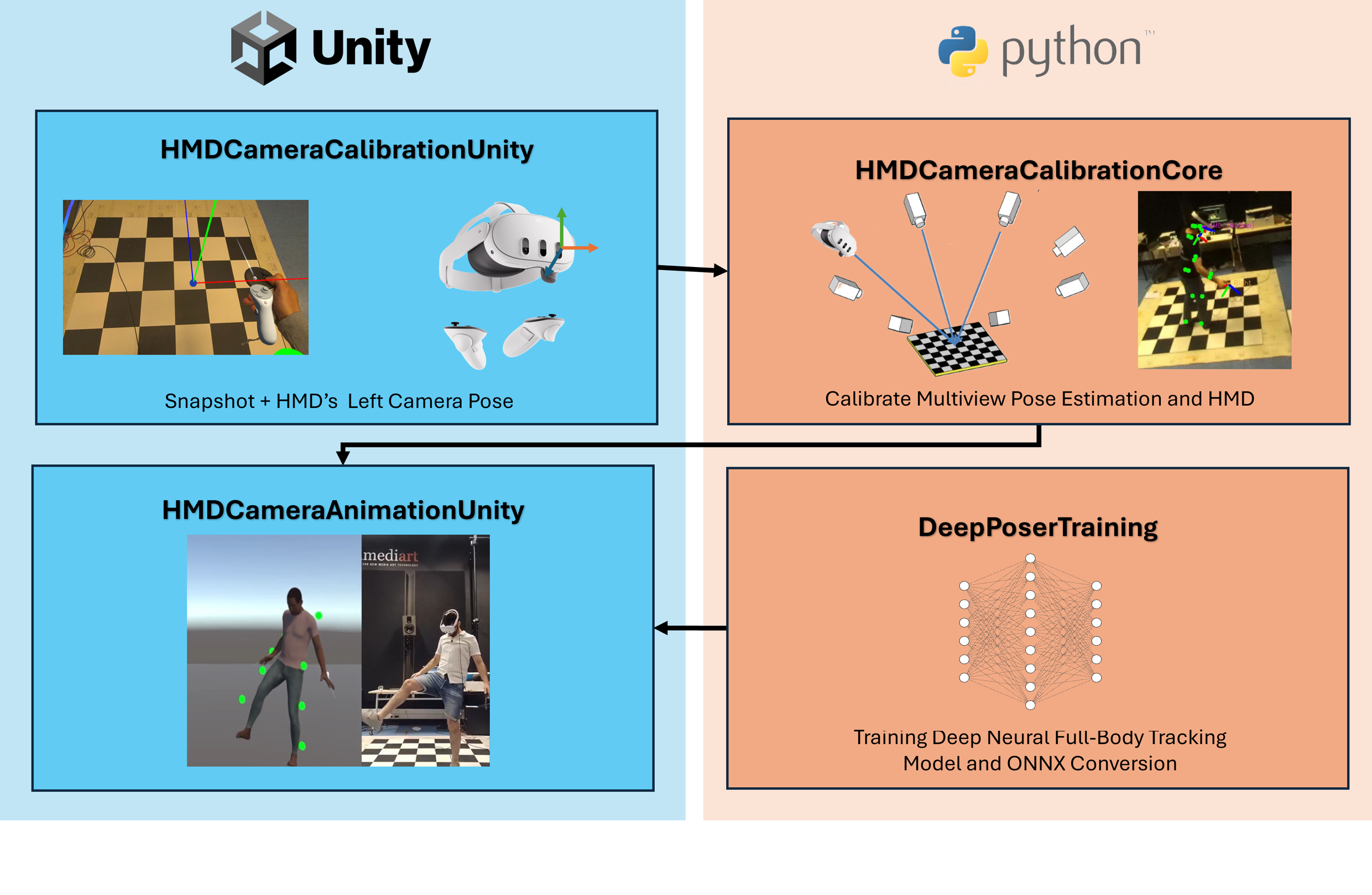}
  \caption{Overview of the proposed framework for self-avatar full-body tracking from an HMD and external RGB cameras. First, the HMD and cameras are calibrated into a common coordinate system, so the 18-DoF HMD and controller motion data and the 3D keypoints from video-based pose estimation are expressed in the same frame. Then, a deep learning-based model takes these sparse tracking signals and 3D keypoints as input and estimates the full-body pose. The pipeline runs end-to-end at 60 FPS and supports a variable number of cameras, calibrated into a common coordinate system regardless of their placement.}
  \label{fig:teaser}
}

\abstract{
Accurate full-body tracking in Virtual Reality (VR) is heavily underconstrained when relying solely on consumer-grade sparse sensors, such as Head-Mounted Displays (HMDs) and handheld controllers. Indeed, the self-avatar full-body pose must be inferred based on the 3 available upper-body joints tracked by these devices. This leads to significant tracking error, especially in lower-body pose. To overcome this issue, we propose to extend the tracking setup with external multiview RGB cameras. We show that integrating RGB-based pose estimation this framework is able to significantly decrease lower-body tracking errors, even under common artifacts in RGB-based estimates. Additionally, we evaluate a only-once HMD-camera calibration procedure that does not require any physical markers. We finally propose an implementation of the overall framework, achieving an end-to-end animation at 60 FPS on a Nvidia RTX3080 GPU. Codes are available at \url{https://github.com/antmaio/MuCHMDPoser}.}


\begin{document}

%% The ``\maketitle'' command must be the first command after the
%% ``\begin{document}'' command. It prepares and prints the title block.

%% the only exception to this rule is the \firstsection command
\firstsection{Introduction}

\maketitle
In Virtual Reality (VR), a key component of immersive experience is the concept of the self-avatar \ie a digital representation that embodies the user within the virtual environment. A fundamental requirement for effective self-avatars is their ability to accurately follow the user's real-world motions with high fidelity. Many works highlight the importance of an accurate full-body tracking to maintain user experience, especially regarding user-centered qualitative metrics such as Sense-of-Embodiment (SoE) or Sense-of-Agency (SoA) \cite{fribourg_avatar_2020,yun_animation_2023}.

To track user's motion, immersive applications often rely on consumer grade VR hardware \ie Head-Mounted Display (HMD) and handheld controllers. These devices integrate IMU-based trackers to accurately capture head and hands motion. In this context, animating the full body of the articulated self-avatar is not straightforward. The system must infer the complete body pose from the motion signals of the head and hands. This problematic is even more challenging regarding lower-body pose reconstruction. Indeed, this configuration does not involve any lower-body tracking as input to animate the lower-body of the self-avatar with an high fidelity.

To address these limitations, various approaches incorporate additional tracking modalities. Some methods use supplementary IMUs or wearable sensors attached to the lower body \cite{ponton_sparseposer_2023,ponton_dragposer_2025}, while others leverage RGB cameras \cite{yang_hybridtrak_2022,pan_fusing_2023,fletcher_improving_2023,maiorca_robustness_2025}. Camera-based systems offer a non-intrusive alternative: they require no user-worn hardware and avoid per-user calibration. 

%However, standard 2D pose estimation algorithms, particularly fast, off-the-shelf models, often produce inaccurate or occluded joint detections. In multi-view setups, these errors propagate into 3D triangulation, producing jitter in the lifted 3D poses. Consequently, these artifacts in the reconstructed 3D positions degrade the quality and fidelity of the full-body reconstruction \cite{maiorca_self-avatar_2024}. 

In this context, we propose a framework for fast full-body pose estimation in large-scale VR environments. Our setup relies on multiple RGB cameras strategically placed around the scene to capture additional motion cues, particularly for the lower body, helping to resolve pose ambiguities that arise when only head and hand signals are available. The resulting 3D joint positions are then integrated with the VR motion features to estimate the full-body avatar pose.

% \begin{itemize}
%     \item We introduce a deep learning-based model that estimates the full-body pose from the 18DoF HMD and controller motion data and the 3D keypoints obtained from RGB-based pose estimation, and we show that it significantly reduces lower-body tracking errors, even under common artifacts in RGB-based estimates.
%     \item We evaluate a one-time, marker-free HMD-camera calibration procedure that expresses both modalities in a common coordinate system, and that supports a variable number of cameras.
%     \item We propose an implementation of the complete pipeline that achieves end-to-end animation at 60~FPS on an Nvidia RTX3080 GPU.
% \end{itemize}
\section{Related Work}

\subsection{Full-body Pose Estimation from Sparse IMUs Tracking in VR}
\label{subsec:vr}
Estimating the full-body pose of an articulated avatar from 18-DoF motion data is a major challenge in Extended Reality (XR), primarily due to the complete absence of lower-body tracking in consumer-grade XR hardware. Early solutions relying on Inverse Kinematics \cite{caserman_analysis_2019}, Machine Learning \cite{ahuja_coolmoves_2021} or Motion Matching \cite{aiide_2016_motion_2016,ponton_combining_2022} have increasingly been superseded by Deep Learning techniques. 

Modern generative techniques such as diffusion \cite{castillo_bodiffusion_2023,dittadi_full-body_2021,du_avatars_2023} and normalizing flows \cite{aliakbarian_flag_2022} have been introduced to synthesize more natural motion by modeling the underlying data distribution. Building on this generative philosophy, Starke et al. recently introduced codebook matching techniques capable of generating full-body motion across diverse scenarios, such as sports or dance, while enabling character control via traditional inputs like joysticks \cite{starke_categorical_2024}.

Sequence modeling approaches are now widely used to capture temporal and spatial dependencies and predict poses end-to-end \cite{jiang_avatarposer_2022,zheng_realistic_2023,yang_lobstr_2021,di_dual_2023,ponton_sparseposer_2023}. Following this trend, \textit{HMD-Poser} further optimizes this by employing a lightweight temporal-spatial network for on-device inference, while incorporating online body shape estimation to improve accuracy across diverse user morphologies \cite{dai_hmd-poser_2024}. To further address the computational bottlenecks of Transformer-based architectures, \textit{SSD-Poser} introduces a State Space Attention Encoder, leveraging State Space Duality to achieve high-precision pose reconstruction with significantly lower latency and a smaller memory footprint \cite{zhao_ssd-poser_2025}. 

%\orange{Also talk about SAGE and highlight limitations!}

While efficient, these techniques often rely on reliable hand tracking, making it difficult to adapt when motion signals are lost. Egocentric video-based methods have been introduced to tackle this problem by modeling hand visibility within the HMD field-of-view \cite{aliakbarian_hmd-nemo_2023,jiang_egoposer_2025}. Similarly, \textit{ReliaAvatar} employs an auto-regressive architecture to ensure robustness against both short- and long-term tracking signal loss \cite{qian_reliaavatar_2024}, while \textit{DragPoser} utilizes a variational latent space to maintain pose sampling efficiency and adherence to real motion distributions during such outages \cite{ponton_dragposer_2025}. 

\subsection{Multimodal Full-body Pose Estimation in VR}

Despite recent advancements, a common limitation shared across methods presented in Section \ref{subsec:vr} is the reduced accuracy of synthesized lower-body motion compared to upper-body gestures. Furthermore, even the most recent models struggle to generalize across a wide variety of static poses due to inherent ambiguities in estimating lower-body configurations from sparse input signals \cite{starke_categorical_2024}.

To address the inherent ambiguity of lower-body pose estimation, several full-body animation approaches rely on complementary sensing modalities to recover additional motion cues. Optical motion capture systems provide highly accurate motion measurements at high frame rates and have been successfully used in VR contexts \cite{wu_marker-based_2025,ha_full-body_2022}. However, these systems require expensive infrastructure and cumbersome body-worn markers or sensors, which limits their scalability and practical deployment outside controlled environments. Consequently, more affordable body-tracking solutions have been proposed that combine inertial measurement units (IMUs) with additional sensing technologies. In particular, integrating IMUs with Ultra-Wideband (UWB) localization has been shown to mitigate the drift and jitter inherent to inertial tracking \cite{armani_ultra_2024,liu_umotion_2025,xue_group_2025}. Despite improving robustness, these approaches still rely on body-worn hardware and typically operate with a single instrumented user per sensor setup. Furthermore, they require per-user calibration procedures to align the different sensing modalities.
In contrast, vision-based approaches using RGB cameras enable passive motion capture without body-worn hardware, and require only a one-time extrinsic calibration rather than per-user calibration. Several works fuse IMU measurements with 2D pose estimates from RGB images to alleviate occlusion and depth ambiguity \cite{pan_fusing_2023,yang_hybridtrak_2022,bao_fusepose_2022,yang_posevr_2024}. \textit{FusePose} \cite{bao_fusepose_2022} combines inertial and visual observations through kinematic modeling and state-space fusion, while VR-oriented systems such as \textit{HybridTrak} \cite{yang_hybridtrak_2022} and \textit{PoseVR} \cite{yang_posevr_2024} augment headset and controller tracking with additional RGB views to estimate lower-body motion. However, these approaches are typically designed around single-view configurations, which limits their scalability to larger environments and multi-user scenes.

Camera-based systems are, moreover, restricted to the space they cover. To overcome this constraint, recent works estimate full-body motion from head-mounted sensors only, such as egocentric cameras and IMUs, leveraging stereo or temporal visual cues \cite{li_egoposeformer_2026,leonardis_egoposeformer_2025,tran_head2body_nodate}. Yet, egocentric views suffer from severe self-occlusion and a narrow field of view that often leaves the lower body out of sight, making the kinematics of invisible joints highly ambiguous. These systems are also prone to ``false positives,'' where limbs of other people are attributed to the user.

Observing the user from the outside avoids these issues. Multiview setups reduce self-occlusion, since a joint hidden in one view usually remains visible in another, and per-view 2D detections can be lifted to 3D by triangulation, typically through the Direct Linear Transform (DLT) \cite{hartley_multiple_2004}. Yet, external cameras alone are not sufficient to animate a self-avatar: their 3D estimates remain noisy and are expressed in the camera frame rather than in the HMD tracking space, whereas the HMD and controllers already provide accurate, low-latency head and hand measurements. The two modalities are therefore complementary.

Building on this, we propose a framework that fuses sparse HMD and controller signals with 3D keypoints from external multiview RGB cameras. Our contributions are:
\begin{itemize}
    \item A deep learning-based model that estimates the full-body pose from 18-DoF HMD and controller data and RGB-based 3D keypoints, significantly reducing lower-body tracking errors, even under common artifacts in RGB-based estimates.
    \item A one-time, marker-free HMD-camera calibration procedure that expresses both modalities in a common coordinate system and supports a variable number of cameras.
    \item An implementation of the complete pipeline achieving end-to-end animation at 60~FPS on an Nvidia RTX3080 GPU.
\end{itemize}

\section{Full-Body Pose Estimation}

\subsection{Problem Formulation}

This work addresses the challenge of accurately animating full-body self-avatars using sparse, multimodal body tracking for immersive applications. The core objective is to define a mapping function $f_\phi$ that transforms motion data from consumer-grade VR hardware $\mathbf{m}_{J_{VR}}^{t}$ and off-the-shelf RGB video-based multiview pose estimations $p^t_{J_{ext}}$ into an articulated full-body avatar that follows accurately user's movements. The overview of this framework is shown in Figure \ref{fig:teaser}. 

We assume that head and hands nodes $J_{VR} = \{HMD, LH, RH\}$ are tracked by consumer-grade level devices \ie HMD and handheld controllers. For each timestep $t$, we collect motion features $\mathbf{m}^t_{J_{VR}}  = (m^t_{HMD}, m^t_{LH}, m^t_{RH})$, respectively, for head, left and right hand node. we define these motion features as global position $p^t_{J_{VR}} \in \mathbb{R}^{|J_{VR}|\times 3} $ and rotation $\mathbf{R}^t_{J_{VR}} \mathbb{R}^{|J_{VR}| \times 6}$, as well as their respective linear and angular velocities, $v^t_{J_{VR}} \in \mathbb{R}^{|J_{VR}| \times 3}$ and $\omega^t_{J_{VR}} \in \mathbb{R}^{|J_{VR}| \times 6}$. Rotations are expressed in 6D, particularly useful in deep learning due to the continous mapping of this representation in Eculidean space \cite{zhou_continuity_2019}. Following previous works \cite{dai_hmd-poser_2024}, these motion features are also expressed in head space coordinate, leading to the set of motion features for HMD and hands defined in Equation \ref{eq:vr_motion_features}. $J_{VR}/\text{h}$ denotes motion features expressed in HMD space. 

\begin{equation}
\label{eq:vr_motion_features}
\mathbf{m}_{J_{VR}}^{t} = \left[ 
    \left\{ p_{J_{VR}}^{t}, \mathbf{R}_{J_{VR}}^{t}, v_{J_{VR}}^{t}, \omega_{J_{VR}}^{t} \right\}, 
    \left\{ p_{J_{VR}/\text{h}}^{t}, \mathbf{R}_{J_{VR}/\text{h}}^{t}, v_{J_{VR}/\text{h}}^{t}, \omega_{J_{VR}/\text{h}}^{t} \right\} 
\right]
\end{equation}

The second motion data stream consists of global 3D Cartesian positions $p^t_{J_{ext}} \in \mathbb{R}^{|J_{ext}|\times 3}$ of body joints $J_{ext}$. These positions are estimated using off-the-shelf, multiview 2D pose estimation on videos combined with a 3D pose lifting method $\Psi$, as stated by Equation \ref{eq:pose_lifter}.

\begin{equation}
    \label{eq:pose_lifter}
    p^t_{J_{ext}} = \Psi \left( \left\{ \left( u^t_{J_{\text{ext}}, c}, w^t_{J_{\text{ext}}, c} \right) \right\}_{c \in \mathcal{C}}, \left\{ s^t_{J_{\text{ext}}, c} \right\}_{c \in \mathcal{C}} \right)
\end{equation}

$\left\{ \left( u^t_{J_{\text{ext}}, c}, w^t_{J_{\text{ext}}, c} \right) \right\}_{c \in \mathcal{C}}, \left\{ s^t_{J_{\text{ext}}, c} \right\}_{c \in \mathcal{C}}$ are respectively 2D positions in image space and the estimation confidence score for joint $j \in J_{ext}$ in camera $c \in \mathcal{C}$, where $\mathcal{C}$ is the set of involved calibrated cameras.
\newline

Hence, this configuration leads to an input vector formed by the concatenation of the sensor data and the estimated 3D positions $\left( \mathbf{m}^t_{J_{VR}},p^t_{J_{ext}} \right)$. The predicted full-body motion of the self-avatar is given by
\begin{equation}
    \left[ \hat{T}^t, \hat{\mathbf{R}}^t_{root}, \hat{\Theta}^t_t \right] = f_\phi\left( \mathbf{m}^t_{J_{VR}},p^t_{J_{ext}} \right) 
\end{equation}
where $\hat{T}_{root}^t$ and $\hat{\mathbf{R}}^t_{root}$ represent the predicted global body displacement and orientation, while $\hat{\Theta}^t$ denotes the predicted local orientation for each bone.

% The system learns a mapping 
% $f_\phi:\mathcal{X}\rightarrow \mathcal{Y}$. In standard configuration \cite{du_avatars_2023,jiang_avatarposer_2022,zheng_realistic_2023}, input space $\mathcal{X}$ consists of 18DoF motion signals from HMD and handheld controllers.

\subsection{Input Data Acquisition}

\begin{figure*}
    \centering
    \includegraphics[width=0.65\linewidth]{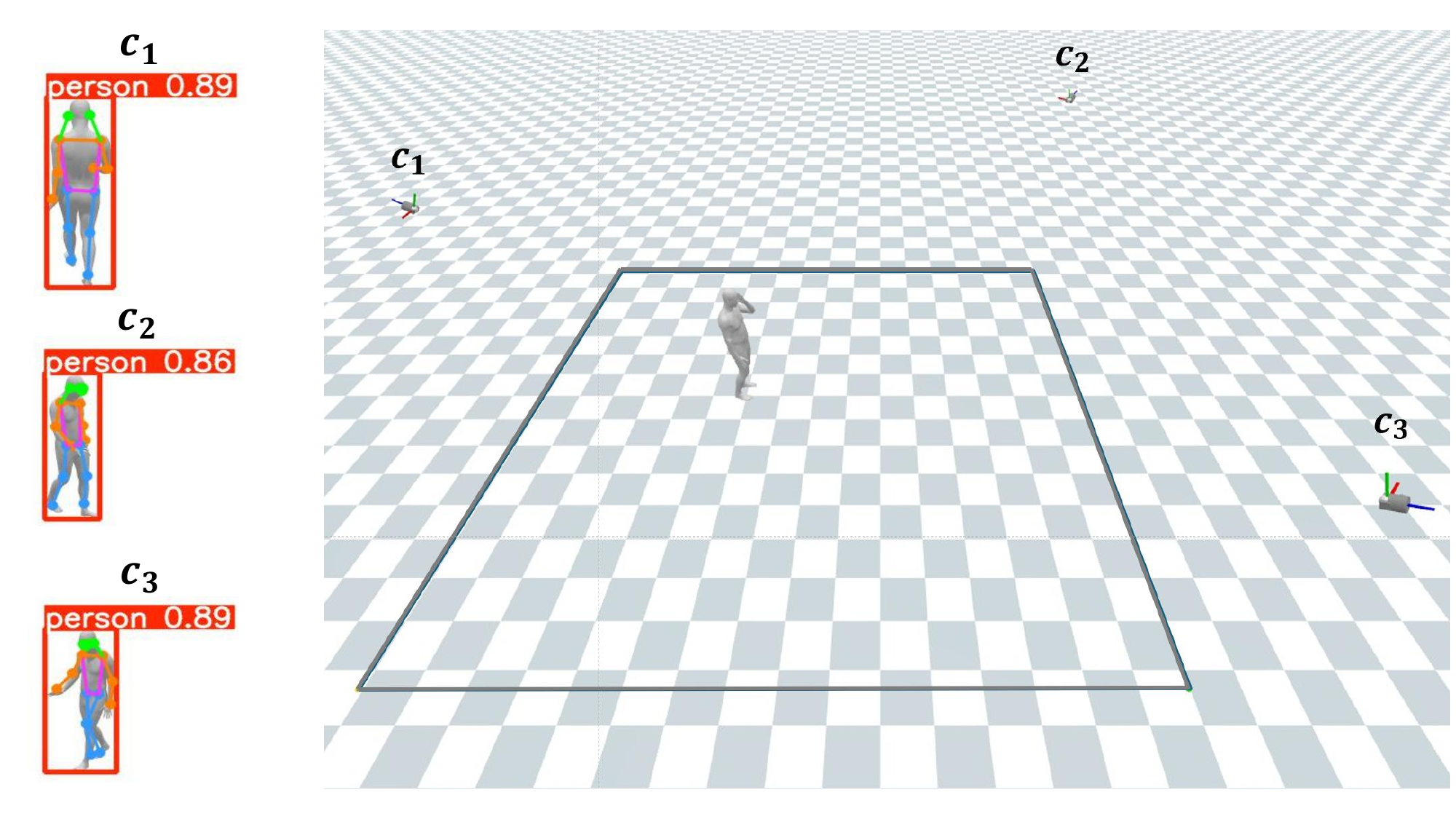}
    \caption{Virtual configuration that imitates hardware setup. We define 3 virtual cameras, whose the intrinsic and extrinsic parameters are known. Off-the-shelf 2D pose estimation is applied in every rendered view to capture lower-body motion cues.}
    \label{fig:scene}
\end{figure*}

% \begin{figure*}
%     \centering
%     \includegraphics[width=0.65\linewidth]{figures/scene.pdf}
%     \caption{Virtual scene for multiview video rendering. Cameras are strategically located to capture the whole range of motion samples defined within the square on the ground.}
%     \label{fig:scene}
% \end{figure*}

We use AMASS \cite{mahmood_amass_2019} as ground truth motion data source. It unifies multiple optical motion capture datasets via the SMPL body model \cite{loper_smpl_2015}. From this dataset, we extract head and hands motion $\left\{ p^t_{J_{VR}}, \mathbf{R}^t_{J_{VR}}, v^t_{J_{VR}}, \omega^t_{J_{VR}} \right\}$ to simulate the tracking signals from a HMD and controllers.
%AMASS provides a unified representation of multiple optical marker-based motion capture datasets by employing the SMPL parametric body model \cite{loper_smpl_2015}.

%It unifies several optical marker-based motion capture datasets through common SMPL representation \cite{loper_smpl_2015} to parametrize human body. 

\begin{figure}[htbp]
    \centering
    \includegraphics[width=0.9\columnwidth]{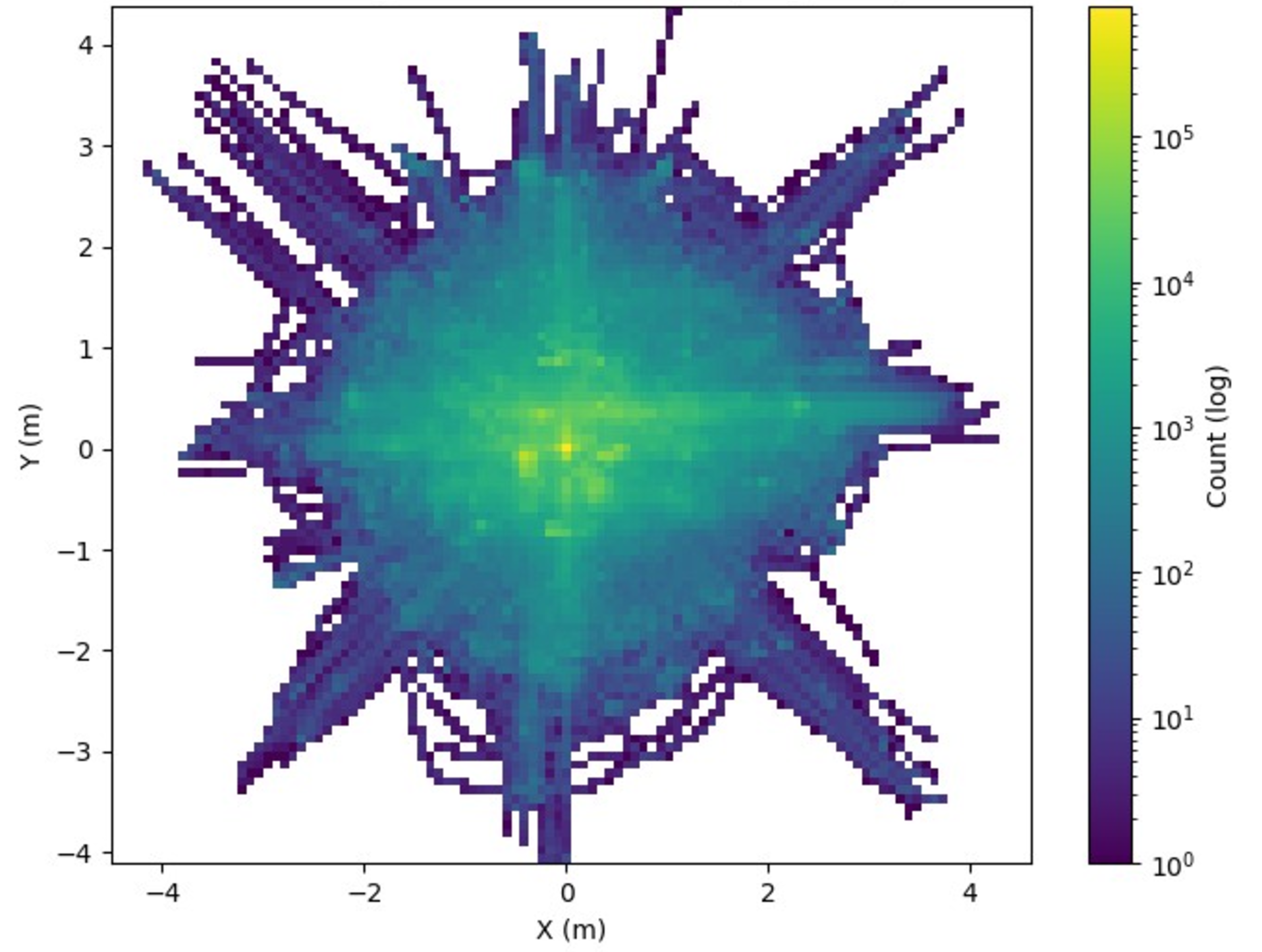}
    \caption{Root global position distribution on the ground surface under a standard evaluation protocol using AMASS dataset. This delimitates an approximative $60m^2$ interaction area.}
    \label{fig:root}
\end{figure}

The 3D Cartesian coordinates of key body joints $p_{J_{ext}}^t$ are collected through off-the-shelf multiview 2D pose estimations on videos rendering of the motion dataset. These videos are rendered from virtual cameras strategically positioned in the scene. A sample of the scene rendering with the virtual cameras $c \in \mathcal{C}$ is shown as matter of example in Figure \ref{fig:scene}. The extrinsic parameters of these virtual cameras are optimized to ensure the entire range of motion within the dataset remains within the field of view. The distribution of the root positions on the ground plane within the dataset is displayed in Figure \ref{fig:root}. Moreover, the cameras intrinsic parameters are fixed to match the specifications of a real-world ZED2 camera setup.

From these rendered videos, we extract 2D joint keypoints position $\left( u^t_{J_{\text{ext}}, c}, w^t_{J_{\text{ext}}, c} \right)_{c \in \mathcal{C}}$ and the related confidence scores $\left( s^t_{J_{\text{ext}}, c} \right)_{c \in \mathcal{C}}$ using an off-the-shelf 2D pose estimation model. A significant challenge is that such models, while fast, are prone to estimation artifacts \eg occlusions or jitter, which severely degrade the quality of 3D poses recovered through standard pose lifting methods. 

To tackle this effect, $\Psi$ from Equation \ref{eq:pose_lifter} combines the \textit{Direct Linear Triangulation} (DLT) method with Kalman filter to estimate smooth 3D positions of full-body joints $p_{J_{ext}}^t$ from the multiple 2D poses estimated in each camera $c \in \mathcal{C}$. For each frame $t$ and joint $j \in J_{ext}$, we apply DLT using the 2D detections from the two cameras with the highest confidence scores $s_{j,c}^t$. If either of these two selected cameras yields a confidence score $s_{j,c}^t < \kappa$, where $\kappa = 0.5$ is our confidence threshold, we consider the joint as occluded and set its 3D position $p_j^t = \mathbf{0}$.

\subsection{Architecture}

\begin{figure}[h]
    \centering
    \hspace*{-15pt} % Shifts the image approximately 15 pixels to the left
    \includegraphics[scale=0.8]{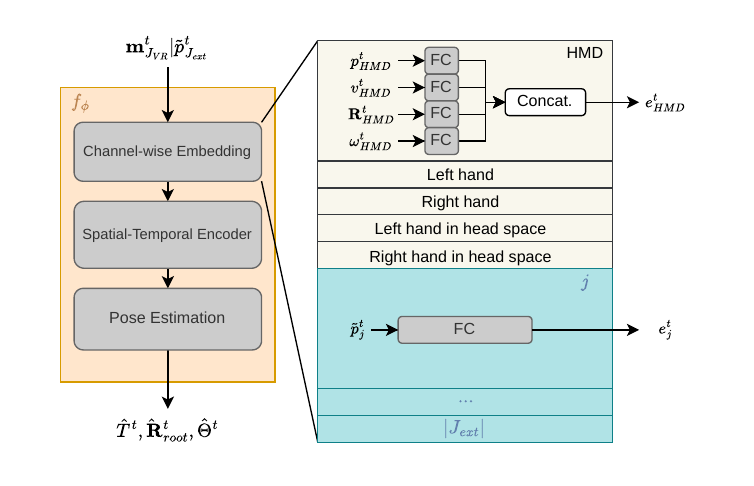}
    \caption{Full-body Pose Estimation model. We extend the architecture from \textit{HMD-Poser} \cite{dai_hmd-poser_2024} to feed the filtered positions $p^t_{J_{ext}}$ of body joints $J_{ext}$ acquired by the multiview pose estimation setup. Extending the channel-wise embedding, we are able to provide embeddings $e_j^t$ from body joints $J_{ext}$ outside standard head and hands tracking.}
    \label{fig:fphi}
\end{figure}

As defined in related work \cite{maiorca_robustness_2025}, we rely on an extended version of \textit{HMD-Poser} \cite{dai_hmd-poser_2024} to estimate the full-body pose of the self-avatar, suitable for its lightweight, yet effective avatar animation framework. The overview of the proposed architecture can be found in Figure \ref{fig:fphi}. First, the motion features from HMD and handheld controllers $\mathbf{m_{J_{VR}}^t}$ are projected into an embedding space in a channel-wise manner. We extend this embedding layer to incorporate 3D positions $p^t_{J_{ext}}$ from the pose estimation framework $\Psi$. Then, the spatial-temporal encoder consists in LSTM and Transformer blocks, to capture respectively the temporal dynamic and spatial relationship in motion data. Finally, we estimate the full state of the articulated avatar \ie global translation and orientation as well as local bone orientation, using a fully-connected-based layer as pose estimation from previoulsy encoded information.     

\begin{figure*}
    \centering
    \includegraphics[width=\linewidth]{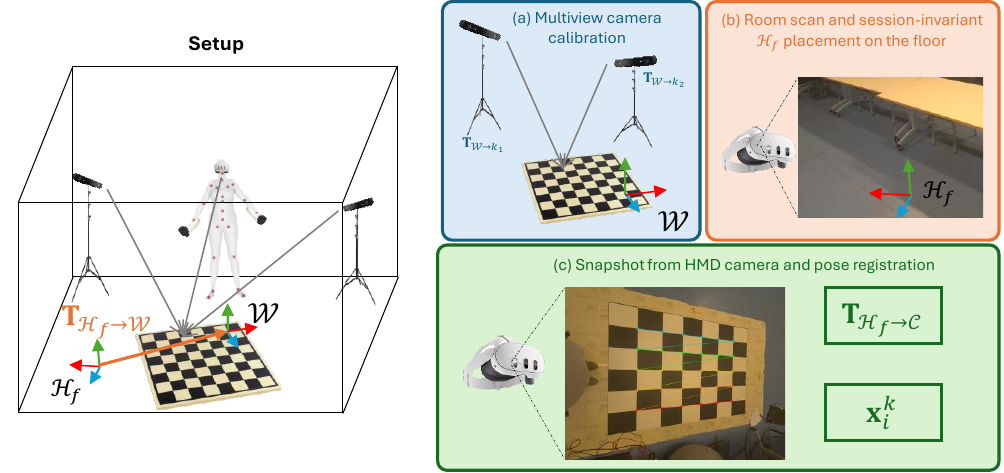}
    \caption{Overview of the calibration setup used to register the multi-camera system, the HMD, and the handheld controllers into a common reference axis system. \textbf{Left}: Schematic of the full setup, showing the subject wearing an HMD inside the capture volume observed by the external RGB cameras. \textbf{Right}: (a) Standard multiview camera calibration protocol \cite{hartley_multiple_2004}. (b) The Meta Mixed Reality Utility Kit (MRUK) provides a room scan from passthrough, allowing a physically-anchored virtual tracking space $\mathcal{H}_f$. (c) Snapshot from the HMD camera $k\equiv\mathcal{C}$ with corner detection $\mathbf{x}_i^k$ and pose registration $\mathbf{T}_{\mathcal{H}_f\to\mathcal{C}}$. This setup closes the loop between the HMD tracking space, the floor-anchored frame within the virtual application, and the external camera system.}
    \label{fig:calibration_setup}
\end{figure*}

\section{HMD and External Camera Rig Calibration}
\label{sec:calibration}

\begin{figure}[htbp]
    \centering
    \begin{tikzpicture}[
        >=Latex,
        thick,
        scale=0.8,
        every node/.style={font=\small}
    ]
        % ---------- Helper macro: coordinate frame with adjustable rotation ----------
        % #1 = origin name, #2 = x-pos, #3 = y-pos, #4 = label, #5 = rotation angle (deg)
        \newcommand{\drawframe}[5]{
            \coordinate (#1) at (#2, #3);
            \begin{scope}[shift={(#2,#3)}, rotate=#5]
                \draw[->, red!70!black]   (0,0) -- (0.8, 0)      node[right=1pt] {\scriptsize $X$};
                \draw[->, green!50!black] (0,0) -- (0, 0.8)      node[above=1pt] {\scriptsize $Y$};
                \draw[->, blue!70!black]  (0,0) -- (-0.45,-0.45) node[below left=-1pt] {\scriptsize $Z$};
            \end{scope}
            \fill[black] (#1) circle (1.2pt);
            \node[above=0.55cm, font=\bfseries] at (#1) {#4};
        }

        % ---------- Domain ensembles (drawn first, so frames sit on top) ----------
        % VR domain: encloses H and Hf
        \draw[dash pattern=on 2pt off 2pt, thick, violet!70!black, rounded corners=8pt]
            (-1.3, -1.9) rectangle (5.3, 1.5);
        \node[font=\bfseries\small, violet!70!black] at (0.7, 1.85) {VR (software)};

        % Physical domain: encloses Hf and W
        \draw[dash pattern=on 2pt off 2pt, thick, orange!80!black, rounded corners=8pt]
            (2.7, -2.1) rectangle (9.3, 1.3);
        \node[font=\bfseries\small, orange!80!black] at (7.3, -2.4) {Physical world};

        % ---------- Place frames with distinct orientations ----------
        \drawframe{H}{0}{0}{$\mathcal{H}$}{0}
        \drawframe{Hf}{4}{0.3}{$\mathcal{H}_f$}{25}
        \drawframe{W}{8}{-0.2}{$\mathcal{W}$}{-15}

        % ---------- Forward transforms ----------
        \draw[->, gray, bend left=25]
            (H) to node[midway, above=0.3cm, sloped, fill=white, inner sep=1pt, font=\footnotesize]
                {$\mathbf{T}_{\mathcal{H}\to\mathcal{H}_f}$} (Hf);
        \draw[->, gray, bend left=25]
            (Hf) to node[midway, above=0.3cm, sloped, fill=white, inner sep=1pt, font=\footnotesize]
                {$\mathbf{T}_{\mathcal{H}_f\to\mathcal{W}}$} (W);

        % ---------- Inverse transforms ----------
        \draw[->, gray, dashed, bend right=25]
            (Hf) to node[midway, below=0.3cm, sloped, fill=white, inner sep=1pt, font=\footnotesize]
                {$\mathbf{T}_{\mathcal{H}_f\to\mathcal{H}}$} (H);
        \draw[->, gray, dashed, bend right=25]
            (W) to node[midway, below=0.3cm, sloped, fill=white, inner sep=1pt, font=\footnotesize]
                {$\mathbf{T}_{\mathcal{W}\to\mathcal{H}_f}$} (Hf);

        % ---------- Compact captions ----------
        \node[align=center, font=\scriptsize\itshape, gray] at (0, -1.3) {session-dependent};
        \node[align=center, font=\scriptsize\itshape, gray] at (4, -1.0) {session-invariant};
        \node[align=center, font=\scriptsize\itshape, gray] at (8, -1.5) {world frame};

    \end{tikzpicture}
    \caption{Reference frames involved in HMD pose estimation. The device-centric frame $\mathcal{H}$, re-initialized arbitrarily at each VR application startup, is mapped to the 
    session-invariant frame $\mathcal{H}_f$. Both $\mathcal{H}$ and $\mathcal{H}_f$ live within the VR software domain, while $\mathcal{H}_f$ additionally serves as the bridge to the 
    physical world domain: the rigid-body transform $\mathbf{T}_{\mathcal{H}_f\to\mathcal{W}}$, obtained from the multi-camera calibration, relates $\mathcal{H}_f$ to the shared physical world frame $\mathcal{W}$.}
    \label{fig:reference_frames_3d}
\end{figure}
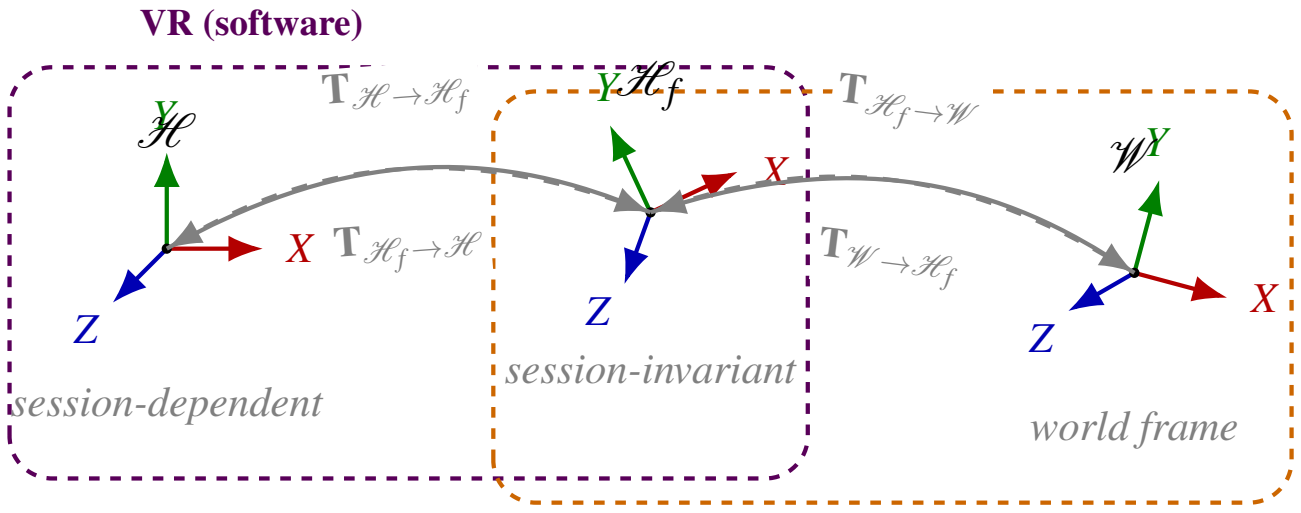

Fusing HMD motion data with video-based 3D keypoints requires both streams to be expressed in a common coordinate system. The two tracking systems, however, are independent. The HMD estimates its own 6-DoF pose through inside-out tracking in a device-centric frame $\mathcal{H}$, whereas the external cameras only know the physical world frame $\mathcal{W}$ defined by their calibration (Fig.~\ref{fig:reference_frames_3d}). Neither system observes the other's frame, so the rigid transform $\mathbf{T}_{\mathcal{H} \rightarrow \mathcal{W}}$ between them must be estimated explicitly. However, this estimation is non-trivial. Indeed, $\mathcal{H}$ is initialized arbitrarily at each application startup, so a transform estimated once is valid for one session only. Then, a session-invariant frame $\mathcal{H}_f$ must therefore be provided by the runtime, allowing a one-time calibration procedure between HMD and cameras. Nevertheless, relocalization errors or tracking failures of $\mathcal{H}_f$ can make it deviate unpredictably. Since every calibrated pose is expressed relative to $\mathcal{H}_f$, such deviations propagate directly into the user's pose in $\mathcal{W}$, without any warning to the application. 

% Reuse the existing TikZ figure (fig:reference_frames_3d), resized to column width.
\subsection{Session-Invariant Reference $\mathcal{H}_f$ Estimation} 

We define $\mathcal{H}_f$ using Meta Mixed Reality Utility Kit (MRUK)\footnote{\url{https://developers.meta.com/horizon/documentation/unity/unity-mr-utility-kit-overview/}}. MRUK provides tools to perform a semantic scan of the room (walls, floor, ceiling, furniture) configured at the OS level before the application starts and exposes the resulting geometry in Unity. We build a deterministic reference axis system $\mathcal{H}_f$ anchored to the ground and aligned with the longest detected wall. The rigid transform $\mathbf{T}_{\mathcal{H}_f \rightarrow \mathcal{H}}  \in SE(3)$ mapping coordinates from axis system $\mathcal{H}_f$ to $\mathcal{H}$ is then known.
%We ask whether this frame is more stable than the raw runtime tracking space $\mathcal{H}$, which serves as our baseline.

\subsection{Calibration approach} Inspired by the standard multiview calibration procedure \cite{hartley_multiple_2004}, we use a planar checkerboard of known geometry as the physical reference shared by both tracking devices. Let $\mathbf{X}_i \in \mathbb{R}^4$, $i=1..N$, be the homogeneous 3D positions of its $N$ inner corners, defined in $\mathcal{W}$ with the board in the $Z=0$ plane. For each external camera $k$, the corners are detected and the extrinsics are recovered by solving the Perspective-$n$-Point (PnP) problem \cite{fischler_random_1981}:
\begin{equation}
    \label{eq:pnp}
    \mathbf{T}_{\mathcal{W}\to k} = \mathop{\mathrm{arg\,min}}_{\mathbf{T} \in SE(3)} \sum_{i=1}^{N} \left\lVert \mathbf{x}_i^{k} - \pi\!\left(\mathbf{K}_k \mathbf{T}\, \mathbf{X}_i\right) \right\rVert^2,
\end{equation}
where $\pi(\cdot)$ denotes the perspective division, $\mathbf{K}_k$ the intrinsics provided by the manufacturer and $\left\{ \mathbf{x}_i^k \right\}_{i=1}^N$ the $N$ corners detected in the image captured by the camera $k$. The transform $\mathbf{T}_{\mathcal{W}\to k}$ represents the camera pose in world reference frame $\mathcal{W}$.  

Then, the HMD is registered with a single snapshot of the same checkerboard, taken with its left passthrough camera $\mathcal{C}$, while we record  the camera pose in $\mathcal{H}_f$, denoted $\mathbf{T}_{\mathcal{H}_f \rightarrow \mathcal{C}}$. The same PnP solve (Equation \ref{eq:pnp}) yields the HMD left camera pose in world reference $\mathbf{T}_{\mathcal{W}\to\mathcal{C}}$. The transform between the virtually and physically anchored reference frames, respectively $\mathcal{H}_f$ and $\mathcal{W}$, is retrieved by transform composition:
\begin{equation}
    \label{eq:intermediate_transform}
    \mathbf{T}_{\mathcal{H}_f\to\mathcal{W}} = \mathbf{T}_{\mathcal{W}\to\mathcal{C}}^{-1}\,\mathbf{T}_{\mathcal{H}_f\to\mathcal{C}}
\end{equation}
Finally, the full pipeline transform between initial tracking space for the 18DoF motion data from consumer-grade VR devices $\mathcal{H}$ and world reference axis system shared across external cameras $\mathcal{W}$
\begin{equation}
        \mathbf{T}_{\mathcal{H}\to\mathcal{W}} = \mathbf{T}_{\mathcal{H}_f\to\mathcal{W}}\,\mathbf{T}_{\mathcal{H}\to\mathcal{H}_f}
\end{equation}

Because $\mathcal{H}_f$ is assumed session-invariant, this registration is performed once. With that procedure, we are then able to express the 6DoF motion data from the HMD in the world reference frame shared with the external RGB cameras. Similarly, we can also express the 3D positions estimated via pose estimation and DLT applied to the camera videos in the virtual tracking spaces $\mathcal{H}_f$ and $\mathcal{H}$.

\section{Experiments}
\label{sec:experiments}

\subsection{Full-Body Pose Estimation}
From RGB video inputs, we estimate the 3D positions of $|J_{ext}| = 12$ body joints, including the feet, ankles, hips, elbows, shoulders, and wrists. The proposed approach is evaluated on the AMASS dataset \cite{mahmood_amass_2019} following two standard evaluation protocols introduced in prior work \cite{zheng_realistic_2023}.

\paragraph{Protocol 1}
This protocol follows a commonly used training and evaluation split constructed from the CMU \cite{carnegie_mellon_university_cmu_nodate}, HDM05 \cite{muller_documentation_nodate}, and BMLrub \cite{troje_decomposing_2002} motion capture subsets. In total, the dataset contains approximately 20 hours of human motion capture data. The training and testing partitions are generated by mixing motion sequences from each subset, following the standard split strategy adopted in recent self-avatar full-body tracking methods \cite{du_avatars_2023,jiang_avatarposer_2022}.

\paragraph{Protocol 2}
In this protocol, the model is trained on a larger set of AMASS subsets, including ACCAD \cite{advanced_computing_center_for_the_arts_and_design_accad_nodate}, BMLrub \cite{troje_decomposing_2002}, BMLmovi \cite{ghorbani_movi_2021}, CMU \cite{carnegie_mellon_university_cmu_nodate}, EKUT \cite{mandery_kit_2015}, Eyes Japan Dataset \cite{eyes_japan_co_ltd_eyes_nodate}, KIT \cite{mandery_kit_2015}, HDM05 \cite{muller_documentation_nodate}, MPI Limits \cite{akhter_pose-conditioned_2015}, SFU \cite{simon_fraser_university_and_national_university_of_singapore_sfu_nodate}, and TotalCapture \cite{trumble_total_2017}. Evaluation is performed on the Transitions Mocap \cite{mahmood_amass_2019} and HumanEva \cite{sigal_humaneva_2010} subsets, which are excluded from training.

Protocol 1 comprises approximately 20 hours of high-fidelity motion capture sequences, whereas Protocol 2 spans roughly 43 hours of recorded motion data. For both protocols, training is performed using a temporal window of 40 frames. We consider both motion streams \ie 3D positions $p_{J_{ext}}^t$ and IMUs trackers $\mathbf{m}_{J_{VR}}^t$ synchronized at constant framerate $60Hz$. The model is trained for 400 epochs on a single RTX 4090 GPU with a batch size of 128. The model is optimized under the same loss function as in related work \cite{dai_hmd-poser_2024}. This optimization is performed using the Adam optimizer with a one-cycle learning rate policy \cite{smith_super-convergence_2018}, reaching a maximum learning rate of $\lambda_{max} = 1e-3$. To estimate 2D positions in each camera view, we rely on the nano model of YOLOv8 framework for a fast data acquisition. 

Finally, full-body pose estimation performance is evaluated through standard objective metrics: Mean Per Joint Positional Error (MPJPE), Mean Per Joint Rotational Error (MPJRE), Mean Per Joint Velocity Error (MPJVE), and the jitter ratio computed between the jitter in predicted and ground-truth global positions of avatar joints.

\subsection{Calibration}

\begin{figure}
    \centering
    \includegraphics[scale=0.8]{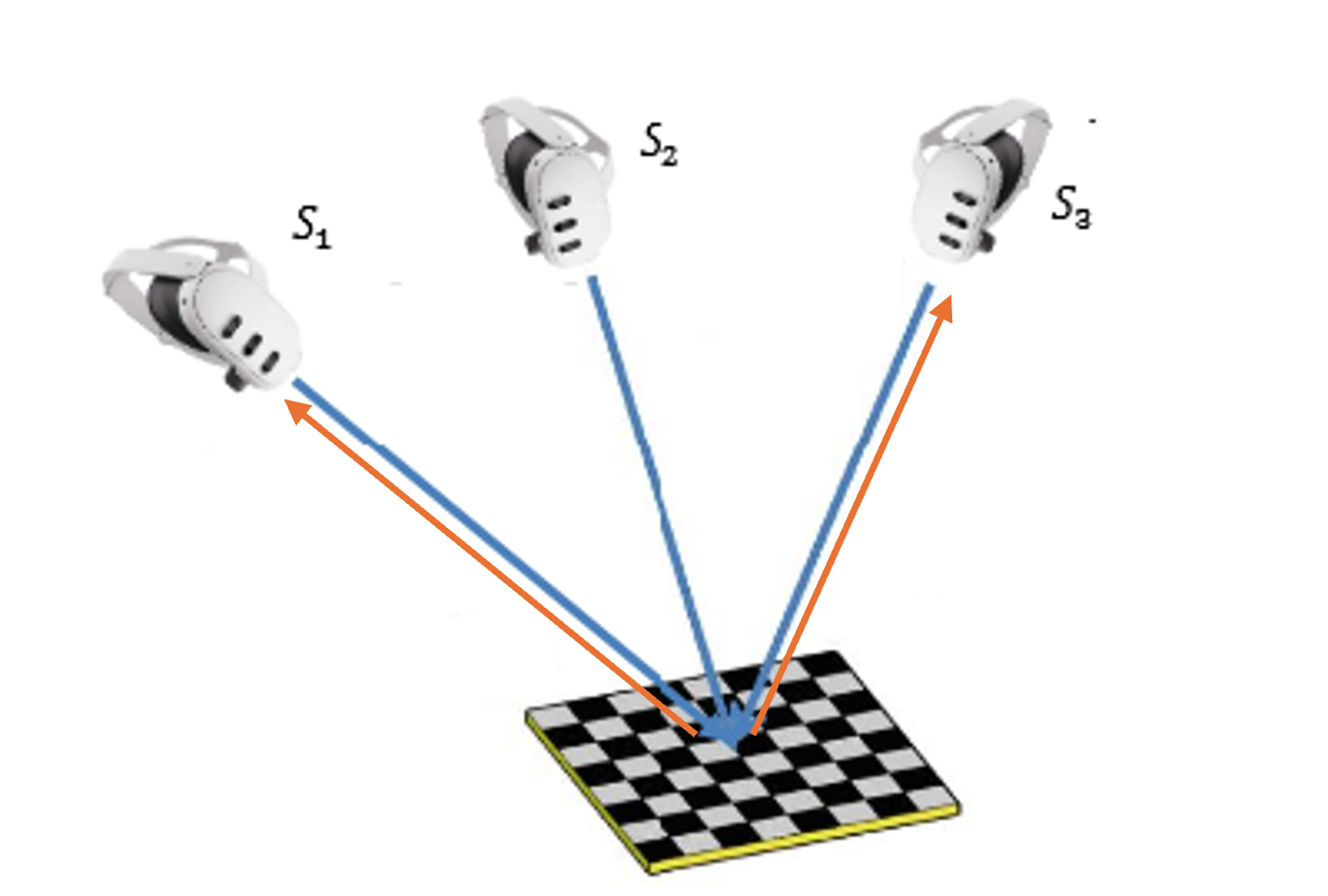}
    \caption{Protocol to assess the consistency and stability of $\mathcal{H}_f$ tracking space. We take several snapshots from left camera of the HMD, register its pose relatively to $\mathcal{H}_f$. We use this transform to reproject the control points into the image plane of each snapshot and measure the discrepancy between these 2D points and the corners detected by OpenCV.} 
    \label{fig:snapshots}
\end{figure}
% \paragraph{Multi-camera calibration.} To evaluate the world frame $\mathcal{W}$ independently of the HMD, we triangulate the four extreme corners of the board from both cameras. These corners form a rectangle of known dimensions, so we measure (a) the side-length consistency $\delta$ between opposite edges and (b) the deviation from orthogonality at each corner. As a pixel-level metric, we reproject the triangulated corners into each camera and report the mean reprojection error $\epsilon_k$ against the detected corners.

To assess the stability of a candidate frame $\mathcal{H}_f$ without external ground-truth equipment, we use a cross-reprojection protocol, shown by Figure \ref{fig:snapshots}. We capture $S = 6$ snapshots $\{\textbf{S}_k\}_{k=1,...,S}$ of the board from different viewpoints, each pairing a passthrough image with the simultaneously recorded camera pose in $\mathcal{H}_f$. For every snapshot $k$, we compute $\mathbf{T}_{\mathcal{W}\to\mathcal{C}}^k$ by Equation \ref{eq:pnp}, so the control points can be expressed in the candidate axis system following Equation \ref{eq:intermediate_transform}. In a round-robin fashion, the points obtained from a reference snapshot $a$ are projected into every other snapshot $b \neq a$ using the pose recorded for $b$. We report two metrics:
\begin{align}
    \epsilon_{\text{2D}}^{a} &= \frac{1}{(S-1)N}\sum_{b\neq a}\sum_{i=1}^{N} \left\lVert \pi\!\left(\mathbf{K}_{\mathcal{C}}\mathbf{T}_{\mathcal{H}_f\to\mathcal{C}}^{b}\,\mathbf{X}_{i,\mathcal{H}_f}^{a}\right) - \mathbf{x}_i^{b}\right\rVert_2, \\
    \epsilon_{\text{3D}}^{a} &= \frac{1}{(S-1)N}\sum_{b\neq a}\sum_{i=1}^{N} \left\lVert \mathbf{X}_{i,\mathcal{H}_f}^{a} - \mathbf{X}_{i,\mathcal{H}_f}^{b}\right\rVert_2,
\end{align}
with $\mathbf{X}_{i,\mathcal{H}_f}^{k}=\mathbf{T}_{\mathcal{W}\to\mathcal{H}_f}^{k}\mathbf{X}_{i,\mathcal{W}}$. 

The two metrics capture complementary aspects of frame quality. $\epsilon_{3D}$ isolates the intrinsic stability of the candidate frame $\mathcal{H}_f$ itself, independently of any camera- or projection-related error. Since the checkerboard is rigid and its corners $\mathbf{X}_{i,\mathcal{W}}$ are fixed and known once and for all from the multiview calibration, any discrepancy between $\mathbf{X}_{i,\mathcal{H}_f}^{a}$ and $\mathbf{X}_{i,\mathcal{H}_f}^{b}$ can only originate from an inconsistency between $\mathbf{T}_{\mathcal{W}\to\mathcal{H}_f}^{a}$ and $\mathbf{T}_{\mathcal{W}\to\mathcal{H}_f}^{b}$. Thus, $\epsilon_{3D}$ directly measures whether the candidate frame's pose relative to the world remains consistent across snapshots, or drifts over time and viewpoint. It is expressed in physical units (mm) and does not depend on external camera intrinsics or on the accuracy of the HMD-to-camera transform.

In contrast, $\epsilon_{2D}$ is an end-to-end, application-facing metric. It evaluates whether a 3D point expressed in $\mathcal{H}_f$ from a reference snapshot $a$ correctly reprojects onto the corresponding checkerboard corner when using the camera pose $\mathbf{T}_{\mathcal{H}_f\to\mathcal{C}}^{b}$ recorded in a different snapshot $b$. Unlike $\epsilon_{3D}$, this metric additionally incorporates the accuracy of the HMD's own tracking relative to $\mathcal{H}_f$, the camera intrinsics $\mathbf{K}_{\mathcal{C}}$, and the corner-detection noise in $\mathbf{x}_i^{b}$. Expressed in pixels, it answers a practical question for augmented-reality applications: if virtual content is anchored in $\mathcal{H}_f$ during one session or viewpoint, how many pixels of misalignment will be observed when revisiting it from another viewpoint?

Reporting both metrics allows the source of error to be localized: a low $\epsilon_{3D}$ combined with a high $\epsilon_{2D}$ indicates that the candidate frame itself is stable but that HMD tracking or intrinsic calibration is the limiting factor, whereas a high $\epsilon_{3D}$ indicates that the frame itself drifts, which necessarily propagates into $\epsilon_{2D}$ as well. This comparison is particularly relevant when contrasting the MRUK-based frame (\textit{sceneSpace}) with the baseline $\mathcal{H}$ (\textit{trackingSpace}), as it reveals whether room-scan-based floor anchoring reduces drift ($\epsilon_{3D}$) relative to the native tracking origin, and whether this improvement translates into a user-perceptible reduction in reprojection error ($\epsilon_{2D}$). Note that $\epsilon_{3D}$ and $\epsilon_{2D}$ are reported in different units (mm and pixels, respectively) and are therefore not directly comparable in magnitude, only in relative trend across the two candidate frames.

As hardware tracking devices, we use two ZED2 as external RGB cameras (1280\.px$\times$720.px), capturing videos from their left lens. We employ a checkerboard with $5\times 6$ inner corners and 25\,cm squares. We use MetaQuest 3 as consumer-grade VR devices, and the HMD calibration relies on its left passthrough camera (1280.px$\times$1280.px).  

% Figure: hmd_calibration_snapshot.pdf (caption as in the chapter, label fig:hmd_calibration_snpshot).

\section{Results}
\label{sec:results}

\subsection{Calibration} Table~\ref{tab:cross_reproj} reports the cross-reprojection error $\epsilon_{\text{2D}}$ and the 3D consistency error $\epsilon_{\text{3D}}$ for each reference snapshot. The MRUK-based frame $\mathcal{H}_f$ and the baseline tracking space $\mathcal{H}$ yield identical results, with $\Delta = 0.00$ for every snapshot. Across snapshots, $\epsilon_{\text{2D}}$ ranges from 2.37 to 3.86\,px (mean 3.19\,px) and $\epsilon_{\text{3D}}$ from 0.62 to 1.03\,cm (mean 0.81\,cm). Defining $\mathcal{H}_f$ from the room scan therefore provides the same stability as the raw tracking frame, and both remain subject to the drift of the headset's visual-inertial odometry. This residual drift, below 1\,cm on average, is smaller than the error typically reported for video-based 3D keypoint estimation, so we consider it compatible with our target application of fusing RGB-based keypoints with HMD data. 

\begin{table}[t]
\centering
\small
\caption{Cross-reprojection errors for each reference snapshot, comparing the standard tracking space $\mathcal{H}$ with the MRUK-based frame $\mathcal{H}_f$. The reprojection procedure results in sub-centimer precision, in average. $\Delta = \epsilon(\mathcal{H}_f) - \epsilon(\mathcal{H})$ indicates that both tracking space axis systems share similar stability properties. Lower is better.}
\label{tab:cross_reproj}
\begin{tabular}{lcccccc}
\toprule
& \multicolumn{3}{c}{$\epsilon_{\text{2D}}^{a}$ (px)} & \multicolumn{3}{c}{$\epsilon_{\text{3D}}^{a}$ (cm)} \\
\cmidrule(lr){2-4} \cmidrule(lr){5-7}
Ref. snapshot $a$ & $\mathcal{H}$ & $\mathcal{H}_f$ & $\Delta$ & $\mathcal{H}$ & $\mathcal{H}_f$ & $\Delta$ \\
\midrule
$\mathbf{S}_1$ & 2.37 & 2.37 & 0.00 & 0.62 & 0.62 & 0.00 \\
$\mathbf{S}_2$ & 3.55 & 3.55 & 0.00 & 0.82 & 0.82 & 0.00 \\
$\mathbf{S}_3$ & 3.58 & 3.58 & 0.00 & 1.03 & 1.03 & 0.00 \\
$\mathbf{S}_4$ & 2.82 & 2.82 & 0.00 & 0.77 & 0.77 & 0.00 \\
$\mathbf{S}_5$ & 3.86 & 3.86 & 0.00 & 0.91 & 0.91 & 0.00 \\
$\mathbf{S}_6$ & 2.93 & 2.93 & 0.00 & 0.71 & 0.71 & 0.00 \\
\midrule
Mean & 3.19 & 3.19 & 0.00 & 0.81 & 0.81 & 0.00 \\
\bottomrule
\end{tabular}
\end{table}

% Figures: pnp_cross_reprojection_report.png (regenerate with only trackingSpace and sceneSpace_LongestWall), scene_and_act.png.

% \paragraph{Limitations.} The evaluation was conducted in a single room, with snapshots captured within the same application run, so it measures stability across viewpoints rather than across sessions. Results also depend on lighting, room geometry, SDK version and headset hardware, and the closed-source MRUK implementation prevents diagnosing the root causes of the remaining drift. Finally, the achieved reprojection accuracy suffices for keypoint fusion but not for applications such as 3D scanning, which require sub-pixel consistency.

\subsection{Multiview 2D to 3D positions lifting}

Table~\ref{tab:2D3D} reports the reconstruction quality of direct linear triangulation (DLT), which we use as a fast, lightweight baseline for multiview 2D-to-3D lifting. For a fair comparison, we remove frames where one occlusion occurs for at least one joint, as we zeroed the positions of occluded joints. DLT requires no learned component and runs per frame, which makes it suitable for building the 3D pose data at scale. Its output is nevertheless imperfect: the MPJPE is 12.94\,cm on Protocol~1 and 21.31\,cm on Protocol~2, and the velocity error (155.78 and 206.80\,cm/s) and jitter ratio (51.12 and 72.53) show that the reconstructed trajectories contain frame-to-frame noise. We do not attempt to remove these inaccuracies at the reconstruction stage. Since the DLT poses are part of the training set, the avatar animation network $\Phi$ is exposed to this noise during training and learns to produce smooth, plausible motion from it.

% requires \usepackage{booktabs}
\begin{table}[t]
\centering
\small
\caption{Multiview 2D-to-3D pose lifting with direct linear triangulation (DLT), evaluated on the two protocols. $\downarrow$: lower is better.}
\label{tab:2D3D}
\begin{tabular}{@{}lcccc@{}}
\toprule
\textbf{Protocol} & \textbf{Occlusion} & \textbf{MPJPE} $\downarrow$ & \textbf{MPJVE} $\downarrow$ & \textbf{Jitter} $\downarrow$ \\
& \textbf{rate [\%]} & \textbf{[cm]} & \textbf{[cm/s]} & \textbf{ratio} \\
\midrule
Protocol 1 & 0.95 & 12.94 & 155.78 & 51.12 \\
Protocol 2 & 0.58 & 21.31 & 206.80 & 72.53 \\
\bottomrule
\end{tabular}
\end{table}

\begin{table*}[t]
\centering
\small
\setlength{\tabcolsep}{7pt}
\caption{Full-body pose estimation accuracy on both protocols, comparing the 3-tracker baseline (HMD and two controllers) with our method, which adds external 3D pose observations. Position errors are reported for the full body (MPJPE), upper body (U-PE) and lower body (L-PE), together with the joint rotation error (MPJRE), joint velocity error (MPJVE) and jitter ratio. $\downarrow$: lower is better; best results in bold.}
\label{tab:merged_metrics}
\begin{tabular}{@{}lcccccc@{}}
\toprule
& \multicolumn{3}{c}{\textbf{Position error}} & \textbf{Rotation} & \textbf{Velocity} & \textbf{Smoothness} \\
\cmidrule(lr){2-4} \cmidrule(lr){5-5} \cmidrule(lr){6-6} \cmidrule(l){7-7}
\textbf{Configuration} & \textbf{MPJPE} $\downarrow$ & \textbf{U-PE} $\downarrow$ & \textbf{L-PE} $\downarrow$ & \textbf{MPJRE} $\downarrow$ & \textbf{MPJVE} $\downarrow$ & \textbf{Jitter} $\downarrow$ \\
& [cm] & [cm] & [cm] & [$^\circ$] & [cm/s] & ratio \\
\midrule
\multicolumn{7}{@{}l}{\textit{Protocol 1}} \\
\addlinespace[0.3em]
3 trackers              & 3.67          & 1.57          & 6.71          & 5.43          & 17.79          & \textbf{1.74} \\
3 trackers + 3D pose (ours) & \textbf{2.35} & \textbf{1.23} & \textbf{3.97} & \textbf{4.62} & \textbf{16.20} & 2.40 \\
\midrule
\multicolumn{7}{@{}l}{\textit{Protocol 2}} \\
\addlinespace[0.3em]
3 trackers              & 5.28          & 2.19          & 9.73          & 9.45          & 26.60          & \textbf{2.10} \\
3 trackers + 3D pose (ours) & \textbf{3.26} & \textbf{1.73} & \textbf{5.48} & \textbf{8.12} & \textbf{23.73} & 3.44 \\
\bottomrule
\end{tabular}
\end{table*}

\subsection{Full-body Pose Estimation}

\begin{figure*}[!h]
    \centering
    \includegraphics[width=0.9\linewidth]{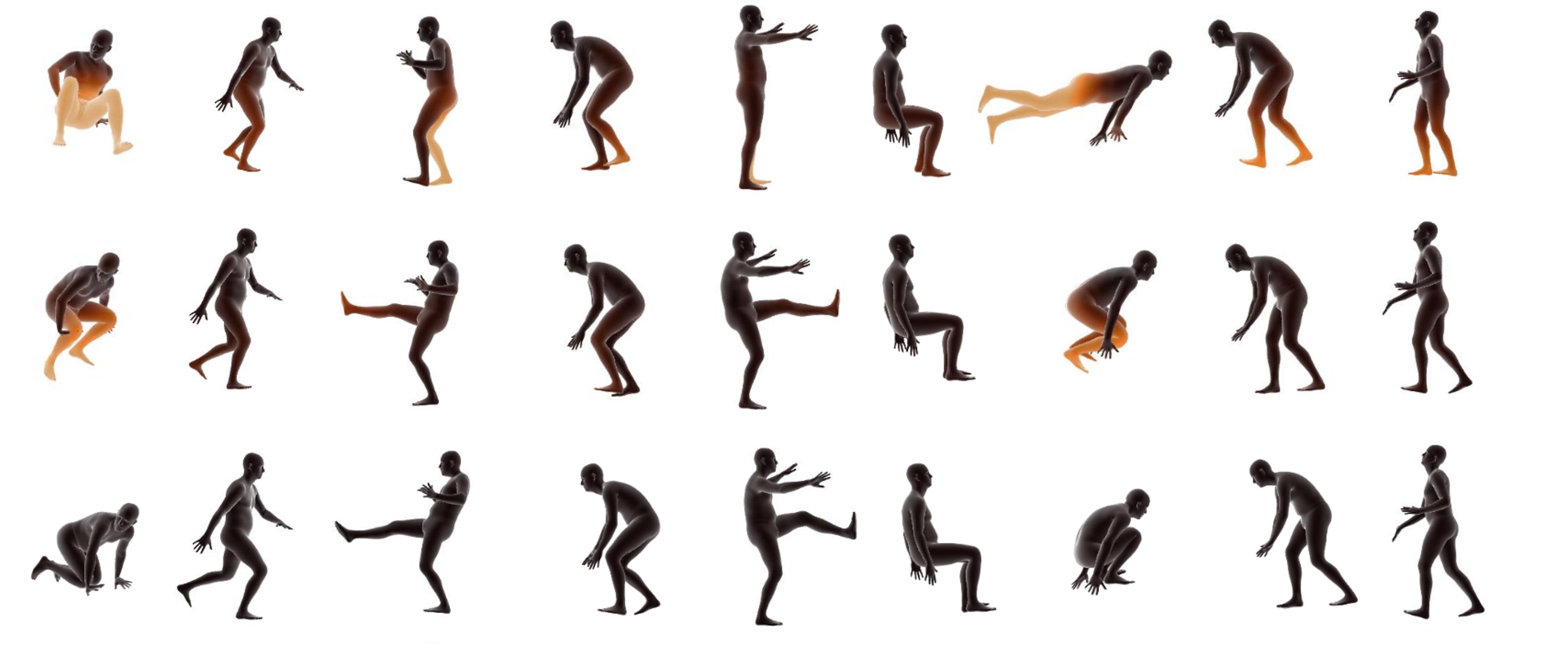}
    \caption{Samples of full-body pose estimated from (\textbf{top to bottom}): 3 trackers; 3 trackers + Pose estimation; ground truth. Incorporating 3D positions from external tracking improves the of the full-body pose estimation, especially at lower-body parts.}
    \label{fig:poses}
\end{figure*}
 
Table~\ref{tab:merged_metrics} presents the quantitative evaluation of our proposed full-body pose estimation pipeline across Protocol~1 and Protocol~2. We compare our fused approach against a baseline relying exclusively on the three standard sparse trackers (the HMD and two handheld controllers). This baseline is similar to the 3-tracker configuration in \textit{HMD-Poser}~\cite{dai_hmd-poser_2024}. We provide samples of generated poses with ground truth in Figure~\ref{fig:poses}.

The 3-tracker baseline exhibits significant ambiguity for untracked joints, with the lower-body position error (L-PE) reaching 9.73\,cm in the dynamic Protocol~2. Integrating external 3D pose observations markedly improves spatial accuracy: MPJPE drops by 36\,\% in Protocol~1, while L-PE is reduced to 5.48\,cm in Protocol~2. Beyond positional accuracy, the fusion also refines rotational and temporal dynamics. MPJRE and MPJVE decrease substantially across both protocols, indicating a more anatomically and kinematically plausible pose reconstruction. The upper-body position error (U-PE) sees more marginal gains, as it is already highly constrained by the HMD and controller hardware. Overall, the improvement across all metrics highlights the effectiveness of the external observations in resolving the degrees of freedom that sparse tracking alone cannot capture.

The external 3D observations also introduce some high-frequency noise into the kinematic chain: the jitter ratio increases from 1.74 to 2.40 in Protocol~1, and from 2.10 to 3.44 in Protocol~2. This increase remains moderate compared with the gains in spatial and rotational accuracy.

\begin{figure*}
    \centering
    \includegraphics[scale=0.27]{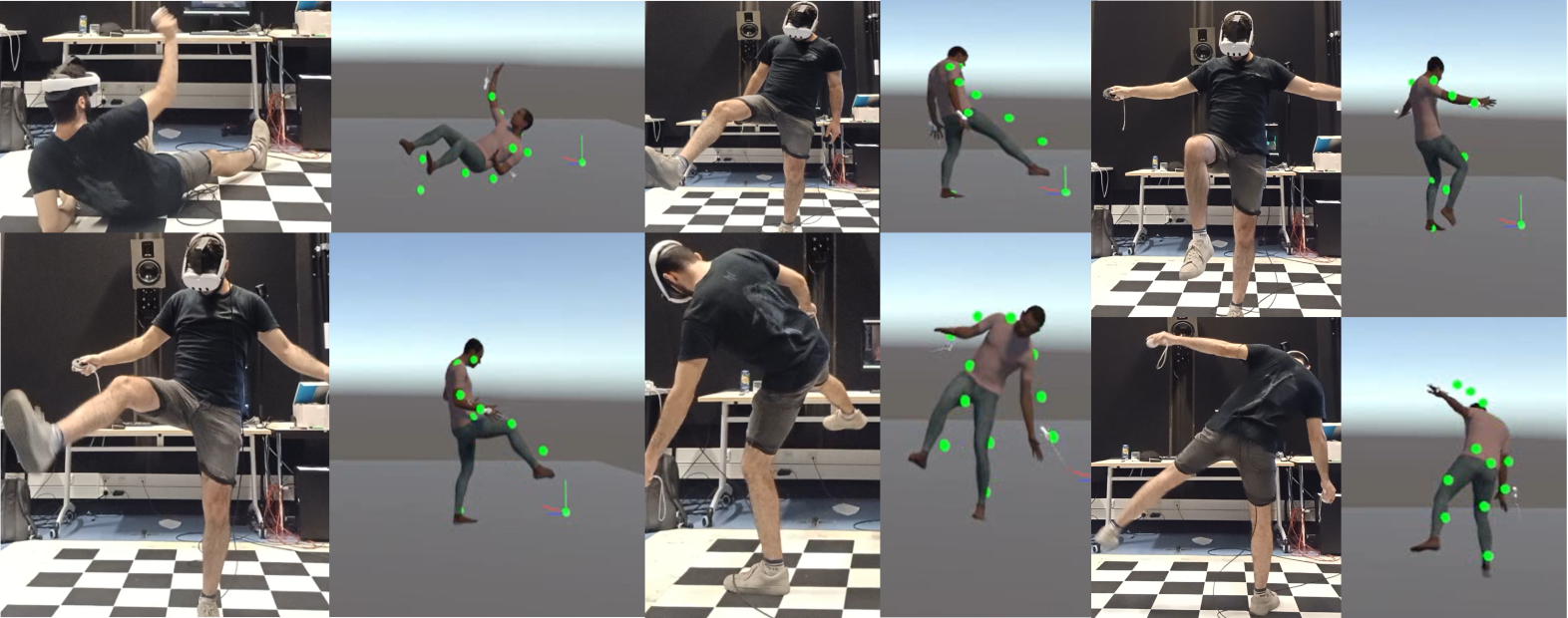}
    \noindent\rule{\linewidth}{0.4pt}\\[15pt]
    \includegraphics[scale=0.35]{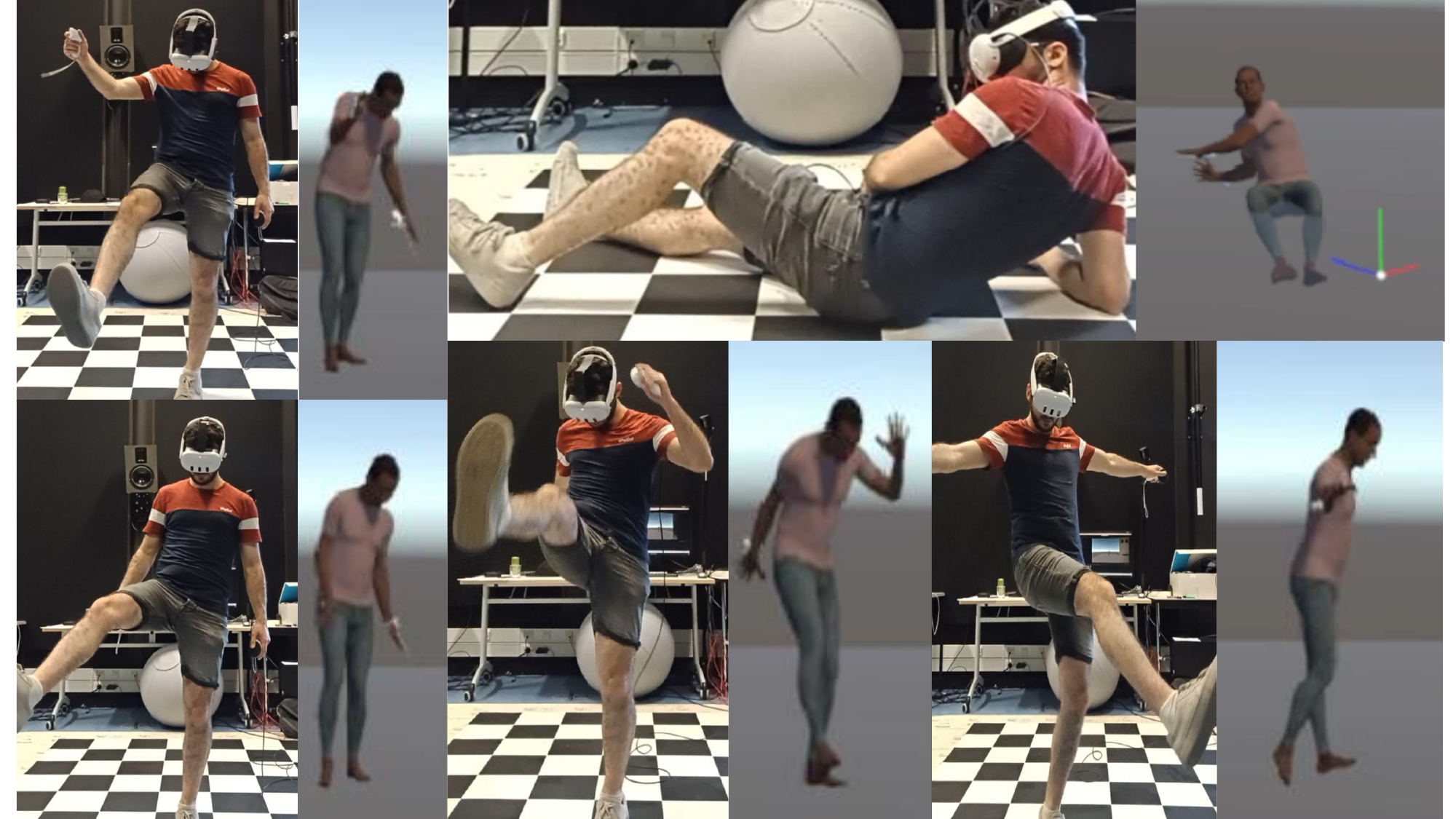}
    \caption{Poses estimated by our proposed framework (\textbf{top}) and from baseline, relying on HMD and controllers solely (\textbf{bottom}). The green spheres represent the positions $p^t_{J_{ext}}$ of body joints estimated from RGB videos (except for the axis system attached to the green sphere, which corresponds to $\mathcal{H}_f$). Our model is able to follow the user's motion and improves lower-body tracking compared to full-body pose estimation from sparse tracking alone.}
    \label{fig:final_output}
\end{figure*}

Figure \ref{fig:final_output} shows sample of self-avatar poses estimated by our proposed framework implemented in game engine. Extending sparse head and hands tracking with RGB video-based tracking enhances the accuracy of the avatar full-body pose, especially at lower-body parts. The inference of our full-body pose estimation framework provides 60FPS animation using Unity Inference Engine \footnote{\url{https://docs.unity3d.com/Packages/com.unity.ai.inference@2.2/manual/index.html}} on a Nividia RTX3080.

% \subsection{Occlusion Robustness}

% We measure joint-wisely the impact of the $v$- and $a$- Kalman filter under occlusion. We flag a frame as occluded when at least one joint is occluded, regarding the confidence threshold definition. We compute the jitter on global joint positions of the avatar on valid frames (no occlusion) and on occluded frames.  
% Figure \ref{fig:robustness_comparison_3way} highlights the efficiency of Kalman Filter to reduce jitter, even under occlusion. 

% \begin{figure*}[htbp]
%     \centering
%     \includegraphics[width=\linewidth]{figures/robustness_comparison_3way.pdf}
%     \caption{Difference of jitter ratio between valid and occluded frames. both Kalman filter implementations help to reduce the negative impact of jitter under occlusion, enabling smoother full-body pose estimation.} 
%     \label{fig:robustness_comparison_3way}
% \end{figure*}

\section{Limitations and Future Work}

Our quantitative evaluation relies on AMASS sequences rendered from virtual cameras with ZED2 intrinsics (Section~3.2), which lets us study full-body accuracy against reliable ground truth. The real-world setup, in contrast, is only assessed through the calibration metrics and qualitative results (Figure~9), since we have no synchronized ground-truth lower-body motion for these recordings. Off-the-shelf 2D detectors may behave differently on real video than on rendered avatars (lighting, clothing, motion blur), which can create a domain gap. Future work should record a real multiview dataset with synchronized optical motion capture, so that real-world pose accuracy can be measured directly.

A second limitation concerns the quality of the 3D observations. We use direct linear triangulation from the two most confident cameras as a fast, lightweight lifting step, and we zero the position of occluded joints. As Table~2 shows, the resulting 3D positions are imprecise, with an MPJPE of 12.94 to 21.31\,cm, and the avatar network is trained on this noisy input. This choice explains part of the jitter increase observed in Table~3 (1.74 to 2.40 in Protocol~1 and 2.10 to 3.44 in Protocol~2), even though spatial and rotational accuracy improve. More robust lifting, such as confidence-weighted triangulation over all available views or a learned multiview lifting module, could provide cleaner inputs at a modest additional cost. The influence of the number and placement of cameras on accuracy also remains to be studied.

The calibration has its own limits. The cross-reprojection protocol (Section~5.2) is run within a single session and shows a residual drift of 0.81\,cm on average, which comes from the headset's visual-inertial odometry. In this setting, the MRUK-based frame $\mathcal{H}_f$ and the native tracking space $\mathcal{H}$ give identical results, so the benefit of room-scan anchoring is not visible here. Its expected advantage is session invariance, which we assume but have not tested across application restarts, relocalization events, or longer sessions. Our implementation also depends on the Meta Quest 3 and MRUK, and a checkerboard is still needed once to register the HMD with the cameras. Evaluating the frame across sessions and adapting the procedure to other headsets are natural next steps.

Moreover, an important open question is how the remaining jitter and spatial inaccuracies affect user-centered metrics, such as the Sense of Embodiment (SoE)~\cite{roth_construction_2020}. While related work highlights the importance of animation fidelity and control in SoE~\cite{yun_animation_2023,fribourg_avatar_2020}, it remains to be determined whether the accuracy gains of our pipeline outweigh the added jitter for the user. Future work must therefore include subjective user evaluations to validate the perceptual benefit of these hybrid tracking methods.

Finally, providing lower-body 3D positions is compatible with a wide range of 3-point tracking architectures, so recent lightweight state-space models such as \textit{SSD-Poser}~\cite{zhao_ssd-poser_2025} could be extended with this approach. A comprehensive benchmark against other sparse tracking methods, including \textit{DragPoser}~\cite{ponton_dragposer_2025} and \textit{SparsePoser}~\cite{ponton_sparseposer_2023}, would further contextualize our results. In addition, our system currently tracks a single user in a large scene. Extending it to multi-user environments, still without additional wearable lower-body trackers, is a promising direction, and its main challenges are user (re-)identification and fast multi-user pose estimation from RGB videos.

\section{Conclusion}
This paper presented a framework for full-body self-avatar tracking in VR that fuses the sparse signals of an HMD and handheld controllers with 3D joint positions estimated from external multiview RGB cameras. We proposed a one-time calibration procedure that expresses both modalities in a common coordinate system through a single checkerboard snapshot and supports a variable number of cameras. We also introduced a deep learning model that estimates the full-body pose from these inputs, trained on noisy triangulated positions. The calibration shows a residual error of 0.81\,cm on average, which is compatible with fusing RGB-based keypoints with HMD data. On AMASS, external 3D observations reduce the MPJPE by 36\,\% and 38\,\% on Protocols~1 and~2, and the lower-body position error by 41\,\% and 44\,\%, at the cost of a moderate increase in jitter. The complete pipeline runs end-to-end at 60\,FPS on a consumer GPU. These results suggest that combining consumer-grade VR devices with outside-in RGB tracking is a practical way to improve avatar full-body pose estimation in large-scale scenes, and that its perceptual benefit for the user should be examined in future work.

% \section{Supplemental Material Instructions}
% \label{sec:supplement_inst}

% In support of transparent research practices and long-term open science goals, you are encouraged to make your supplemental materials available on a publicly-accessible repository.
% Please describe the available supplemental materials in the \hyperref[sec:supplemental_materials]{Supplemental Materials} section.
% These details could include (1) what materials are available, (2) where they are hosted, and (3) any necessary omissions.

% \section*{Supplemental Materials}
% \label{sec:supplemental_materials}

% Refer to the instructions for this section (\cref{sec:supplement_inst}).
% Below is an example you can follow that includes the actual supplemental material for this template:

% All supplemental materials are available on OSF at \url{https://doi.org/10.17605/OSF.IO/2NBSG}, released under a CC BY 4.0 license.
% In particular, they include (1) Excel files containing the data for and analyses for creating \cref{tab:vis_papers} and \cref{fig:vis_papers}, (2) figure images in multiple formats, and (3) a full version of this paper with all appendices.
% Our other code is intellectual property of a corporation---Starbucks Research---and there is no feasible way to share it publicly.

%\bibliographystyle{abbrv}
\bibliographystyle{abbrv-doi}

\bibliography{references}
\end{document}